# Towards Real-World Efficiency: Domain Randomization in Reinforcement Learning for Pre-Capture of Free-Floating Moving Targets by Autonomous Robots

Bahador Beigomi[1], Zheng H. Zhu[2]

*Abstract—* In this research, we introduce a deep reinforcement learning-based control approach to address the intricate challenge of the robotic pre-grasping phase under microgravity conditions. Leveraging reinforcement learning eliminates the necessity for manual feature design, therefore simplifying the problem and empowering the robot to learn pre-grasping policies through trial and error. Our methodology incorporates an off-policy reinforcement learning framework, employing the soft actor-critic technique to enable the gripper to proficiently approach a free-floating moving object, ensuring optimal pre-grasp success. For effective learning of the pre-grasping approach task, we developed a reward function that offers the agent clear and insightful feedback. Our case study examines a pre-grasping task where a Robotiq 3F gripper is required to navigate towards a free-floating moving target, pursue it, and subsequently position itself at the desired pre-grasp location. We assessed our approach through a series of experiments in both simulated and real-world environments. The source code, along with recordings of real-world robot grasping, is available at Fanuc_Robotiq_Grasp.

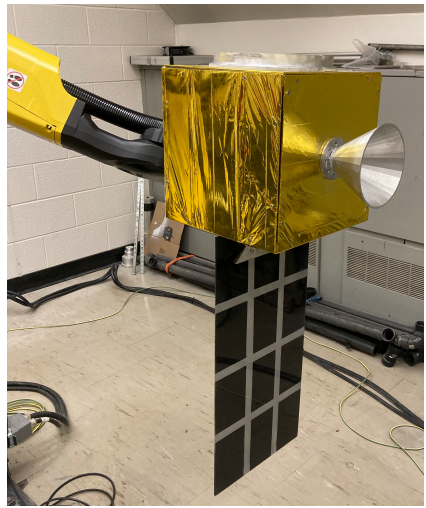

Fig 1. **Free-Floating Target** Manipulation Using a 6-DOF Industrial Robot with Gravity-Compensation for Arbitrary Positioning and Velocity Control

## I. INTRODUCTION

Initial investigations into capturing free-floating objects primarily relied on predictive and offline planning techniques. Most of these methods required prior knowledge of the target's shape, structure, motion trajectory, complex condition-related dynamics equations, and environmental modeling. However, these techniques' effectiveness diminishes when the models lack precision. [1]–[4] As a result, recent years have seen a rise in the exploration of model-free strategies for robotic grasping, aiming to address these limitations. [5]

Conventional grasping studies predominantly focused on manipulating static objects, including both rigid and compliant targets. [6]–[9] Recently, however, there has been an increased demand for advanced autonomous grasping technologies, especially for free-floating entities like lightweight objects affected by wind or space debris in microgravity conditions. The unpredictable environment, the dynamic positioning of moving or tumbling objects, and the risk of pushing these objects away upon contact, present unique challenges. Addressing these challenges demands the development of efficient grasping strategies and real-time interaction policies. Therefore, ongoing research in this area is vital for enhancing robotic systems in dynamic settings.[1]

Deep Reinforcement Learning (DRL) applied to robotics has shown promising results in robotic grasping tasks in recent years, particularly for stationary objects. [10]–[12] Yet, there is a growing need for advanced autonomous grasping technologies capable of handling the challenges posed by free-floating entities. [13] One significant challenge in DRL is bridging the sim-to-real gap, ensuring that real-world robots can effectively operate using policies developed via DRL. One approach to this challenge is randomization, where simulations are randomized during training to produce robust policies. An alternative is the hardware in-the-loop (HIL) process, which entails training directly on the hardware. However, given the slower pace of training on real robots and the potential dangers associated with imperfect policies during training or unforeseen situations, using real robots for direct policy training is generally infeasible. [14]–[16]

Many grasping systems are designed with the assumption of a table-top setting, opting for 3-DoF grasp poses to simplify the complexities of reasoning. [17]–[20] However, when attempting to grasp free-floating target that moves unpredictably in space, an autonomous robot must expand its action space to 6-DoF, as illustrated in Fig 1. Engaging in these environments introduces two prominent challenges: 1) How can a robot reliably follow a free-floating target in motion? and 2) How can it anticipate the impact of tactile sensors on the probability of a successful grasp?

[1]Bahador Beigomi is a PhD candidate at Space Debris Lab, Mechanical Engineering dep., York University, Canada, baha2r@yorku.ca
[2]Zheng H. Zhu is a Professor and Tier I York Research Chair in Space Technology at Department of Mechanical Engineering York University, Canada, gzhu@yorku.ca

*Research supported by CREATE (555425-2021) and (543378-2020) of the Natural Sciences and Engineering Research Council (NSERC) of Canada, and FAST (19FAYORA14) of the Canadian Space Agency.

Motivated by the aforementioned challenges, we propose a DRL-based control system integrated with tactile feedback sensors to position the robotic gripper accurately. This means that the gripper is capable of positioning itself in a way that allows its fingers to surround the target. By doing so, it ensures that if the gripper were to close its fingers, it would successfully grasp the target. This paper offers three primary contributions:

- It presents empirical evidence supporting the feasibility of deriving pre-grasp policies for free-floating moving targets using DRL. This ensures the gripper can be accurately controlled to align itself in the ideal pre-grasp position while tracking the moving free-floating target.

- Real-world evaluation results illustrate that domain randomization in a simulated environment enables the trained agent to bridge the simulation to reality (sim2real) gap, ensuring consistent and reliable actions for robots in real-world scenarios.

- Simulation-based success rates indicate that integrating contact sensing can strengthen the development of pre-grasp policies, enhancing their resilience, especially for lightweight free-floating objects.

These findings underscore our method's potential in paving the way for more dependable and efficient techniques for grasping free-floating moving targets.

## II. TECHNICAL PRELIMINARIES

In this section, we present the principles underlying the DRL framework, delving into the method that is used to control the gripper to place itself in pre-grasp position with respect to a free-floating moving target.

### A. Background

In our study, we examine the application of a DRL framework wherein an agent engages with its environment over discrete time intervals. We model the RL problem using a Markov Decision Process (MDP), a mathematical structure used to represent decision-making across various settings. The MDP is characterized by five elements ($S$, $a$, $\mathcal{P}$, $\mathcal{R}$, $\gamma$): a continuous state space, a continuous action space, the transition dynamics denoting the likelihood of state changes given an action, a reward function which determines the rewards for specific actions in particular states, and a discount factor between 0 and 1 that quantifies the emphasis on future versus immediate rewards. [21]–[23]

The agent aims to formulate a policy ($\pi$) which associates states with action distributions, optimizing the anticipated returns within a given timeframe. In DRL, a neural network, serving as a nonlinear function approximator, often represents this policy. The primary goal is to fine-tune the network's parameters ($\theta$) to ensure optimal performance. [24]

### B. Algorithm

The soft actor-critic (SAC) algorithm [25], used in this work, is a DRL method that integrates an off-policy learning approaches, like Deep Q-Networks (DQN) [26], with actor-critic methodologies such as Proximal Policy Optimization (PPO) [27] and Deep Deterministic Policy Gradient (DDPG) [28]. It uses entropy regularization to balance exploration and exploitation, excelling in robotic control, including motion planning for grasping objects. It speeds learning and avoids local optima by promoting broad exploration through an entropy-based approach within the actor-critic framework. By incorporating entropy into the reinforcement learning objective, it fosters stochasticity and adaptability, ensuring policies are robust across different environments by aiming for maximum entropy in all states:

$$\pi^* = \mathrm{argmax}_\pi \sum_t \mathbb{E}_{(s_t,a_t)\sim\rho_\pi} \gamma^t \left( r(s_t, a_t) + \alpha H(\pi(a_t|s_t)) \right) \quad (1)$$

where the entropy term defined as:

$$H(\pi(a_t|s_t)) = -\sum_{a_t} \pi(a_t|s_t) \log \pi(a_t|s_t) \quad (2)$$

The entropy term, coupled with the scalar temperature parameter ($\alpha$), modulates the balance between exploration (entropy maximization) and exploitation (expected return maximization). A greater temperature parameter emphasizes exploration, whereas a smaller temperature parameter leans towards exploitation. [29] For a given policy, the soft Q-value can be iteratively determined. The soft state value $V(s_t)$ is trained to minimize the mean squared error with estimated soft state value given by:

$$J_V(\psi) = \mathbb{E}_{s_t\sim\mathcal{D}} \left[ \frac{1}{2} \left( V_\psi(a_t) - \mathbb{E}_{a_t\sim\mathcal{D}} \left[ \min_{j=1,2} Q_{\phi_j(s,a)} - \alpha \log \pi_\theta(a_t|s_t) \right] \right)^2 \right] \quad (3)$$

Let $\mathcal{D}$ represent the replay buffer, which stores past experiences including state, action, reward, next state, and termination signal. $\psi$ and $\phi$ respectively denote the parameters of the soft value network and the critic network. This facilitates efficient learning via random sampling. The dual objective soft Q-function for any policy $\pi$ is derived by iteratively applying the modified Bellman backup operator to (1). The parameters of the soft Q-function are trained to reduce the soft Bellman residual.

$$J_Q(\phi) = \mathbb{E}_{(s_t,a_t)\sim\mathcal{D}} \left[ \frac{1}{2} \left( Q_\phi(s_t, a_t) - (r(s_t, a_t) + \gamma \mathbb{E}_{s_{t+1}\sim\rho}[V_\psi(s_{t+1})]) \right)^2 \right] \quad (4)$$

The policy network's objective function utilizes the minimum of these Q-values. To enhance the control model's robustness, we introduce 10% random Gaussian noise ($\xi$) to each action during neural network training to prevent local optima. SAC, rather than defining a deterministic policy, characterizes a policy as a Gaussian distribution. The policy network, for every state, provides the Gaussian's mean and standard deviation, from which an action is derived. This stochastic approach promotes policy exploration. SAC uses the tanh function to ensure actions remain within the [-1,1] range, squashing the Gaussian-sampled action.

$$a_\theta(s,\xi) = \tanh(\mu_\theta(s) + \xi \times \sigma_\theta(s)) \quad (5)$$

where the mean $\mu_\theta$ and the standard deviation $\sigma_\theta$ are output by the policy network $\pi_\theta$, and $\xi$ is the independent noise. To enhance stability in the learning process and expedite convergence, we propose an SAC-based policy training methodology, detailed in Algorithm 1. This approach involves:

- Incorporating entropy regularization in computing both soft state value loss and policy loss.

- Utilizing a target policy network to compute policy loss, merging the clipped double-Q method with target policy smoothing. This aims to prevent Q-function overestimation and reduce suboptimal policy updates.

- Designing a neural network to autonomously adjust the entropy regularization coefficient, ($\alpha$). Elevated values drive exploration in the initial training phases, while diminished values prioritize exploitation during later stages.

- Adopting a linear function to adjust the learning rate of both the actor and critic networks. This progressively reduces the learning rate during training, fostering prompt initial advancements, enhanced parameter space exploration, and more consistent convergence.

## III. PROBLEM FORMULATION

In this section, we discuss the interaction between an agent and its environment within the context of a standard off-policy reinforcement learning framework. We provide a detailed description of the overall task, the observations and actions, and the reward function.

### A. Learning Environment

The designed learning environment leverages the SB3 framework [30], [31] combined with the Pybullet physics engine [32] to emulate the physical system. Pybullet aligns well with RL algorithms and offers efficient and precise simulations, including details on contact points, which is specifically very critical in this work. This simulation context can be observed in Fig 2.

In this research, we chose a flat box as the target object for multiple reasons. Initially, the design of the flat box facilitates the gripper's ability to effortlessly surround the object, ensuring a firm grip. Moreover, the flat box serves as a versatile and representative example of various objects found in practical applications. The dimensions of the box used to train the agent in the simulation environment are $0.2_m \times 0.2_m \times 0.04_m$.

### B. Task Definition

For this study, we introduce an autonomous grasping task, named *Fanuc_Robotiq_Grasp*. This task is challenging for normal RL methods, as it demands tens of thousands of experiences to effectively learn robust state representations and control policies. The task of autonomously grasping a free-floating moving object is characterized by positioning the gripper relative to the target. This task poses significant challenges, especially because one must formulate a reliable policy that addresses the nature of free-floating objects, which can be easily displaced by small forces. The primary aim of the gripper is to accurately align its grip around the moving target and secure it. To assess the ability of our suggested method for mastering this grasp, we utilize the Robotiq 3F gripper. We assume that this gripper is attached to a 6-DOF robotic manipulator, which provides the necessary movement to the gripper's base.

### C. Definition of Actions and States

DRL algorithms for motion planning primarily rely on state and action spaces. Grasping tasks vary greatly between table-top scenarios and targeting free-floating objects, with the latter being more complex. It's crucial to have an observation set that offers detailed process insights and is easily accessible for real-world application. Therefore, action and state space designs must reflect the system's dynamics and challenges.

The coordinates of the moving object and the gripper's pose should both be represented in the inertial frame. This facilitates direct implementation of the trained policy on real robots without fine-tuning. The state space ($S$) encompasses the pose and velocity of the gripper ($G_P, G_v$), the free-floating target's pose and velocity ($T_P, T_v$), relative pose and velocity ($\delta_P, \delta_v$), minimum distance between the target and gripper's palm ($d$), and the total contact force exerted on the target measured by tactile sensors ($F_N$). This results in a 40-dimensional state space vector.

$$S = \{G_P, G_v, T_P, T_v, \delta_P, \delta_v, d, F_N\} \in \mathbb{R}^{40} \qquad (6)$$

The action $a$ is continuous denoting the gripper's movement. It characterizes both its directional and rotational displacements within its own reference frame, defined as follows:

---

**Algorithm1: Policy Training**

**Initialize** soft value $\psi_V$, critics $Q_{\phi_1}, Q_{\phi_2}$, and actor $\pi_\theta$ networks with random parameters $\psi, \phi_1, \phi_2, \theta$

**Initialize** target networks with the same parameters

$\phi_1^{\text{target}} \leftarrow \phi_1$, $\phi_2^{\text{target}} \leftarrow \phi_2$

**Initialize** Replay Buffer $\mathcal{D}$

**for** each iteration **do**

  **for** each step **do**

    Select action with exploration noise $a = \tanh(\mu_\theta(s) + \xi \times \sigma_\theta(s))$

    Store transition tuple $(s, a, r, s', d)$ in $\mathcal{D}$

    Sample mini-batch of N transitions $(s, a, r, s', d)$ from $\mathcal{D}$

  **for** each gradient **do**

    Update soft value network $\psi \leftarrow \psi - lr\, \nabla_\psi J_V(\psi)$

    Update critic networks $\phi_i \leftarrow \phi_i - lr\, \nabla_{\phi_i} J_Q(\phi_i)$

    Update actor network $\theta \leftarrow \theta - lr\, \nabla_\theta J_\pi(\theta)$

    Soft update target critic network $\phi_i^{\text{target}} \leftarrow \tau \phi_i^{\text{target}} + (1-\tau)\phi_i^{\text{target}}$

    Update temperature parameter $\alpha \leftarrow \alpha - lr\, \nabla_\alpha J(\alpha)$

---

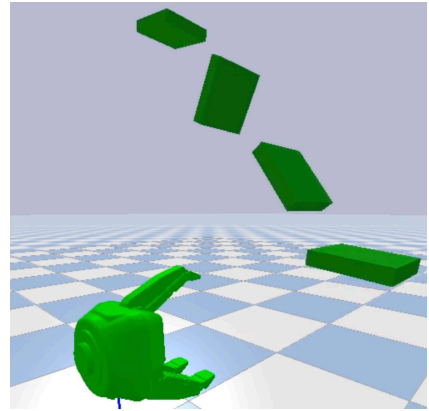

Fig 2. PyBullet environment created for the grasping task of free-floating moving object.

$$a = \{a_x, a_y, a_z, a_\phi, a_\theta, a_\psi\} \in \mathbb{R}^6 \quad (7)$$

At every timestep, the agent can opt to shift by up to 10mm in any given direction and rotate up to 0.1 rad around each axis. The continuous action space enables the gripper to choose any value within the range of -1 to 1, ensuring a uniform action spectrum. Later, the algorithm adjusts the magnitude of the actions for each component. These adjustments are then translated to the manipulator's end-effector, essentially the gripper's base, using the robot's 6dof kinematics.

*D. Reward Shaping*

In RL research, the reward function critically shapes training efficacy. [33] We propose a novel reward function combining dense and sparse rewards to accurately approach and grasp free-floating targets. The agent uses dense rewards for correct alignment with the target, maintaining its position and orientation. Meanwhile, sparse rewards guide optimal posture, ensuring a safe distance from the target and preventing undesired contact. This strategy reduces the chances of missing free-floating moving objects. The reward function is defined as:

$$R = R_d + R_\theta + R_{Top} + P_F \quad (8)$$

To design an optimal reward function, each component is carefully chosen to fall within a range of -1 to 1. This approach offers two primary benefits. Firstly, it removes the need for weight assignment to individual components, as each has an equivalent influence upon activation. Secondly, it simplifies the evaluation and interpretation of training outcomes. The initial two components, $R_d$ and $R_\theta$, provide dense rewards, steering the gripper effectively toward the target.

$$R_d = 1 - \tanh(\delta_d) \quad (9)$$

$$R_\theta = 1 - \tanh(\delta_\theta) \quad (10)$$

$\delta_d$ and $\delta_\theta$ denote the L2 norms for position and orientation differences between the gripper and target, respectively. Both rewards, range from zero to one, with a larger deviation yielding a lower reward. $R_{Top}$ and $P_F$ are sparse rewards and are defined as follows:

$$R_{Top} = \begin{cases} 1 & \frac{n_{in}}{N_p} > 0 \\ 0 & \frac{n_{in}}{N_p} = 0 \end{cases} \quad (11)$$

$$P_F = \begin{cases} 0 & F_N = 0 \\ -1 & F_N > 0 \end{cases} \quad (12)$$

where in $R_{Top}$, depicted in Fig 4, $N_p$ is the total number of object key points, while $n_{in}$ indicates the fraction within the gripper's convex hull, defined by connected adjacent links. The agent earns a positive reward if it positions at least one target key point inside the convex hull. However, any contact between the gripper and the target, showed in Fig 3, measured by $P_F$—the sum of all contact forces detected by tactile sensors—results in a negative reward for the agent. This discourages excessive force, preventing target displacement or potential damage to the gripper. Our study shows that this reward system effectively distinguishes between successful and unsuccessful attempts before their completion.

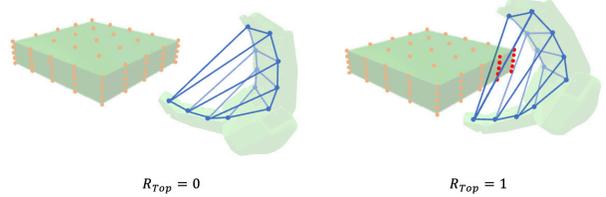

$R_{Top} = 0$      $R_{Top} = 1$
Fig 4 Number of the key points placed in the gripper's convex hull.

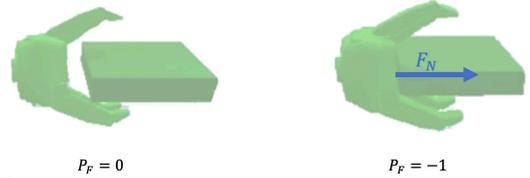

$P_F = 0$      $P_F = -1$
Fig 3. Contact force pushes the target away, making it more likely to miss the grip.

IV. SIMULATION RESULTS

To optimize rewards and train a reliable policy, the robotic gripper must explore all actions in its operational range. Using simulation environments is essential for evaluating various scenarios. Although hardware-in-the-loop setups might produce more reliable outcomes, simulations offer a fast and safe exploration of numerous options. RL algorithms commonly operate under the assumption that a system can be reset after each episode, grant access to observations of all states, and calculate rewards at each timestep, conditions that are often difficult to satisfy in real-world applications. [34]

*A. Training*

The primary objective of this study was to examine the proposed method's ability to identify suitable pre-grasp positions relative to the target and its effectiveness in adapting to new scenarios. For training the policy, we used 40,000 episodes, with each episode consisting of 500 timesteps. At the beginning of each episode, the target is randomly positioned with a constant velocity, which had a maximum relative velocity of $0.4_{m/s}$ with respect to the gripper.

In DRL, the success rate, a metric between zero and one, is widely used to assess an agent's performance. This metric illustrates the consistency with which an agent meets a specific goal within a set environment. In our research, we interpret the success rate as the gripper's ability to achieve more than two positive rewards over 200 consecutive steps. This interval ensures that the gripper can effectively engage in the grasping process. Therefore, the gripper adjusts its position and orientation to maximize dense rewards, as outlined by equations (9) and (10), while positioning a key point within its grasp range without touching the target.

We provide the agent with an enhanced state space and contact data, allowing it to learn relative pose maintenance to the target and position for pre-grasp, all while avoiding unwanted contact that might displace the target. During training, the model saves progress every 40 episodes. The latest best trained agent is then evaluated in five trials to approach a moving free-floating target with random initial states. After 24,000 episodes, training converges as depicted in Fig 5. The model then exhibits a satisfactory average reward and an impressive success rate, with a convergence of 0.91, highlighting its robust learning efficiency in any random situations.

## B. Evaluation

To evaluate the trained agent's performance in a simulated environment, it's crucial to test it across multiple random scenarios and observe how the gripper targets its object. When we load the agent for evaluation, we gather all observations from the simulation. These are then fed into a prediction function, the trained actor network $\theta$, creating a simple input-output process. Using its experience, the agent decides on the best actions, which the gripper then translates into specific movements. Results from the simulation can be seen in Fig 6, covering two unique random episodes.

Both episodes A and B were successful, as they met the criterion for success: maintaining a reward greater than 2.0 for over 200 consecutive timesteps. The initial positions for the targets in episodes A and B were $(0.28, 0.72, 1.05)_m$ and $(-0.29, 0.70, 1.21)_m$, respectively. The corresponding initial velocities were $(-0.21, -0.17, 0.03)_{m/s}$ for episode A and $(0.01, 0.15, 0.14)_{m/s}$ for episode B. While the stationary gripper's starting position was consistent at $(0.0, 0.0, 1.0)_m$ in both episodes, the initial orientations differed: episode A started with an orientation of $(-2.86, -1.02, 2.52)_{rad}$, and episode B with $(-1.71, 0.28, 0.58)_{rad}$. The actions in episode A led to smooth displacements and rotations that effectively guided the agent toward the target. In contrast, the actions in episode B oscillated around what appeared to be the most likely optimal action. This oscillatory behavior is attributable to the training regimen of the agent. As described in Algorithm 1, a 10% noise factor was added to each action chosen by the actor network to enhance policy exploration. These oscillations originated from this added noise, and the agent utilized these fluctuating actions to diminish the effects of noise and maximize its cumulative reward without diverging significantly from its objective. Importantly, as indicated by the data, these oscillations had no impact on the gripper's position or orientation, affecting only its velocity.

In this study, the agent was trained to grasp a free-floating target in motion, with a relative velocity of up to $0.4_{m/s}$ with respect to the gripper. Upon analysis, it was determined that the episodes with the highest rates of failure corresponded to instances where this relative velocity was significantly high. Intuitively, as the velocity of the target increases, the likelihood of the gripper missing the target also rises. To provide a clearer perspective on how variations in speed could influence the results, we refer to TABLE I. , which presents outcomes at different maximum velocities. For the purpose of evaluation, the trained agent controlled the gripper across 100 different random episodes.

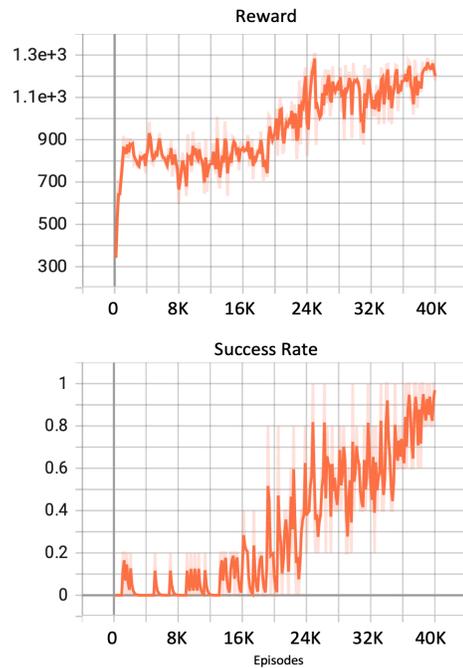

Fig 5. Training Results in Simulation Environment. Mean Evaluation result given best trained agent over 5 random episodes.

Reducing maximum relative velocity generally improves mean reward and success rate, showing a key link to performance. However, randomness in episodes, like initial positions or orientations, may cause some outcome variability, indicating that while the trend is clear, individual results can differ.

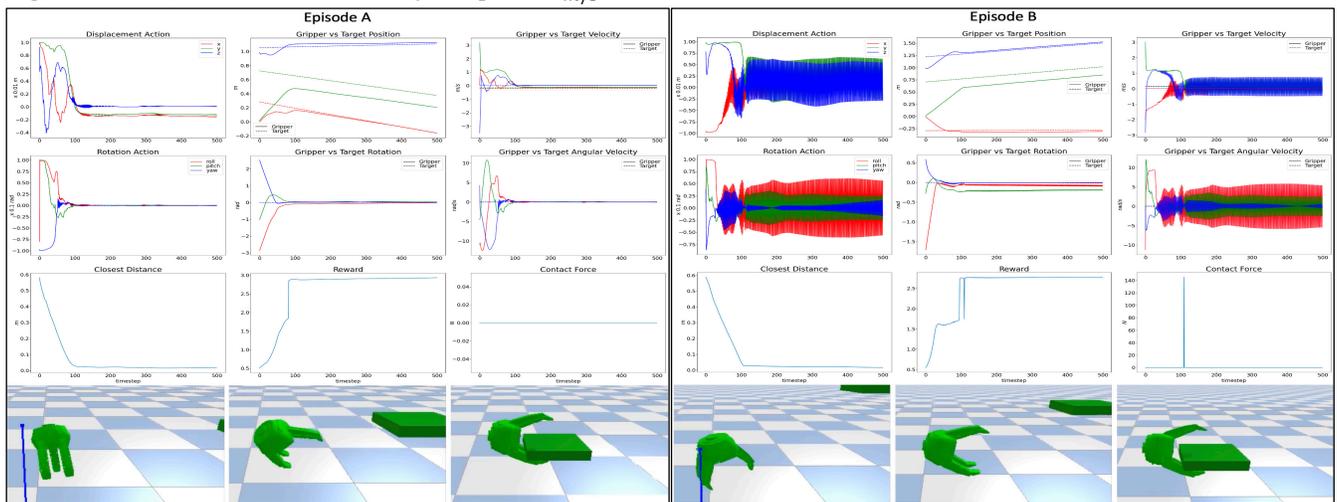

Fig 6. Two distinct successful episodes are depicted. The term 'closest distance' refers to the minimum distance between the target and the gripper's palm, while 'position' indicates the location of their center of mass. In Episode A, the gripper successfully positions itself in the pre-grasp stance without making contact with the free-floating moving target, and it aligns its velocity accordingly. In Episode B, there is slight contact between the target and the gripper's finger; however, the agent manages this effectively.

TABLE I.    EVALUATION FOR DIFFERENT VELOCITIES

| | Maximum Relative Velocity | | | |
|---|---|---|---|---|
| | *0.4 m/s* | *0.3 m/s* | *0.2 m/s* | *0.1 m/s* |
| Mean Eval Reward | 1167.13 | 1208.05 | 1219.57 | 1240.11 |
| Mean Reward Standard Deviation | 131.05 | 185.01 | 103.20 | 80.86 |
| Mean Eval Success Rate | 0.91 | 0.89 | 0.93 | 0.96 |

## V. EXPERIMENTS

In this section, we discuss the experimental results, outline challenges encountered in transitioning from sim2real, and describe the sensors employed. After training a robust agent in the simulation environment with an acceptable success rate, we can practically deploy it on robots to assess its effectiveness in addressing real-world challenges. The process of transferring the trained agent for real-world testing necessitates several considerations:

- With the gripper now affixed to the real manipulator robot, it is unavoidable to utilize the robot's inverse kinematics to determine the subsequent position and orientation of the gripper, when executing the most recent action predicted by the trained agent.

- Contrary to the simulation environment where all observations can be seamlessly obtained from readily available data (e.g., target position or the minimum distance between the target and the gripper's palm), in a real-world setting, these observations must be derived using the available sensors.

- There is a need to adjust the coordination to mirror that of the simulation environment. Furthermore, ascertaining the ground truth position and orientation with respect to an inertial frame for each component becomes paramount when navigating real-world scenarios.

In Fig 7, we illustrate the comprehensive HIL connections that facilitate real-world testing. The setup comprises two primary computer systems, each tasked with controlling different robots. These computers are programmed to communicate with Fanuc M20id robots using UDP (User Datagram Protocol) connections. The primary system, which governs the gripper via a trained agent, randomly selects a position for the target. Subsequently, it dispatches a command to the secondary system, instructing it to relocate the target. Once the system initiates this movement, the primary setup remains in a waiting state until the target attains its predetermined final position.

Upon reaching this state, the closed-loop control of the gripper is activated. The feedback loop sources its positional data from an Intel RealSense camera affixed to the gripper, creating an eye-in-hand configuration. This camera is adept at pinpointing the desired grasping section of the target using the pre-trained YOLOX algorithm. [35] This algorithm is trained to identify a grasping point on an object by creating a bounding box and analyzing it with image processing and depth sensor data. It integrates this with data from robot and tactile sensors, allowing the trained agent to decide the next best action using its simulation training. The system updates the robot's position and orientation by adding the predicted gripper movement each cycle. It uses the robot's Denavit-Hartenberg table and the Levenberg-Marquardt method to calculate the optimal joint angles for the desired motion. [36]

To simplify the process and avoid potential damage when controlling the gripper with a trained agent, we tested the agent exclusively on a stationary free-floating target. The free-floating ability is achieved through the robot's Gravity-Compensation capability, utilizing ATI F/T sensors attached to each robot's end-effector. These sensors calculate external forces applied to the robot and adjust its movements accordingly. Fig 8 presents snapshots of the gripper approaching to grasp the target after it has been randomly positioned in space.

## VI. CONCLUSION

In this paper, we explore the use of DRL for developing pre-grasp policies for free-floating moving objects, demonstrating that the gripper can effectively align its position with and track these targets. Empirical evidence supports the enhancement of policy robustness, especially against lightweight objects, through contact sensing. Our simulations reveal the potential of DRL in learning control maneuvers from experience, confirming the feasibility of training continuous control actions based on state observations within a reasonable timeframe. Domain randomization enhances the trained agent's reliability, effectively bridging the sim2real gap for successful real-world applications without further training. In summary, our work underscores the significance of DRL in creating effective pre-grasp strategies for moving objects, setting the stage for advancements in robotic manipulation.

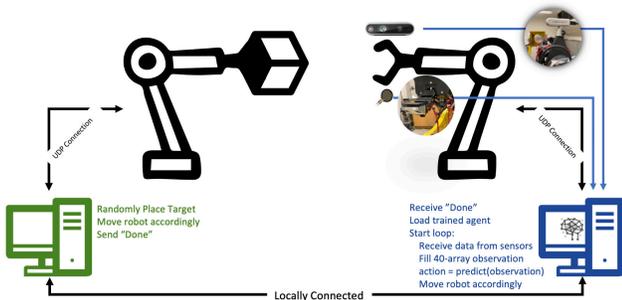

Fig 7. Overall structure of the HIL connection. Each component serves a distinct role in achieving precise control over the gripper's motion.

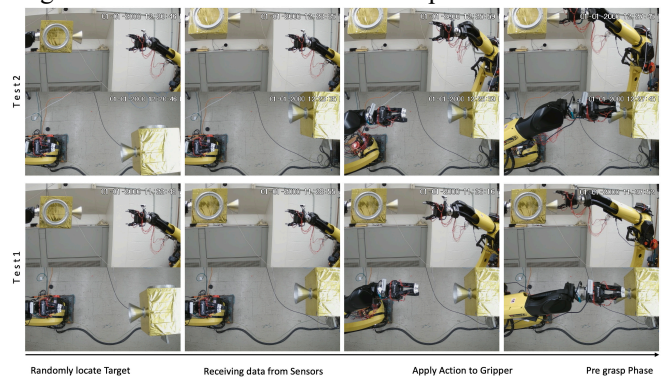

Fig 8. Movement of the Target and Gripper in a Real-world Setting: Upon positioning the target at random location, the gripper aligns itself to a pre-grasp position. Videos of all grasp attempts, both simulation and real-world can be found on the GitHub page.